\documentclass[Afour,sagev,times]{sagej}

\usepackage{moreverb,url, float, subcaption,enumerate}

\usepackage[colorlinks,bookmarksopen,bookmarksnumbered,citecolor=red,urlcolor=red,breaklinks=true]{hyperref}

\newcommand\BibTeX{{\rmfamily B\kern-.05em \textsc{i\kern-.025em b}\kern-.08em
T\kern-.1667em\lower.7ex\hbox{E}\kern-.125emX}}

\def\volumeyear{2019}

\begin{document}

\runninghead{}

\title{Deep learning surrogate models for spatial and visual connectivity}

\author{Sherif Tarabishy\affilnum{1}, Stamatios Psarras\affilnum{1}, Marcin Kosicki\affilnum{1}, Martha Tsigkari\affilnum{1}}

\affiliation{\affilnum{1}Foster + Partners - Applied Research and Development Group}

\corrauth{Sherif Tarabishy}

\email{starabishy@fosterandpartners.com}

\begin{abstract}
Spatial and visual connectivity are important metrics when developing workplace layouts. Calculating those metrics in real-time can be difficult, depending on the size of the floor plan being analysed and the resolution of the analyses.  This paper investigates the possibility of considerably speeding up the outcomes of such computationally intensive simulations by using machine learning to create models capable of identifying the spatial and visual connectivity potential of a space. To that end we present the entire process of investigating different machine learning models and a pipeline for training them on such task, from the incorporation of a bespoke spatial and visual connectivity analysis engine through a distributed computation pipeline, to the process of synthesizing training data and evaluating the performance of different neural networks.
\end{abstract}


\maketitle

\section{Introduction}
\subsection{Motivation}
During the past decade the notion of performance driven design has pushed the industry to develop an array of tools capable of allowing real-time performance metrics to be made available to designers as early as in concept stages. To that end, the response time of analytical tools has been highlighted as a critical aspect of performance-driven design \cite{Chronis2012}. Simulations analysing how well connected a spatial configuration is, in terms of traverse-ability, proximity and visual connectivity, are extremely time-consuming, especially when analysing large office floor plans.

The authors of this article are members of the Applied Research and Development group at Foster + Partners - a global studio for sustainable architecture, urbanism and design. The office’s dedication to performance driven design has been instrumental to the development of tools which allow intuitive yet informed decision making early on in the design process.

\begin{figure}
    \centering
    \includegraphics[width=\linewidth]{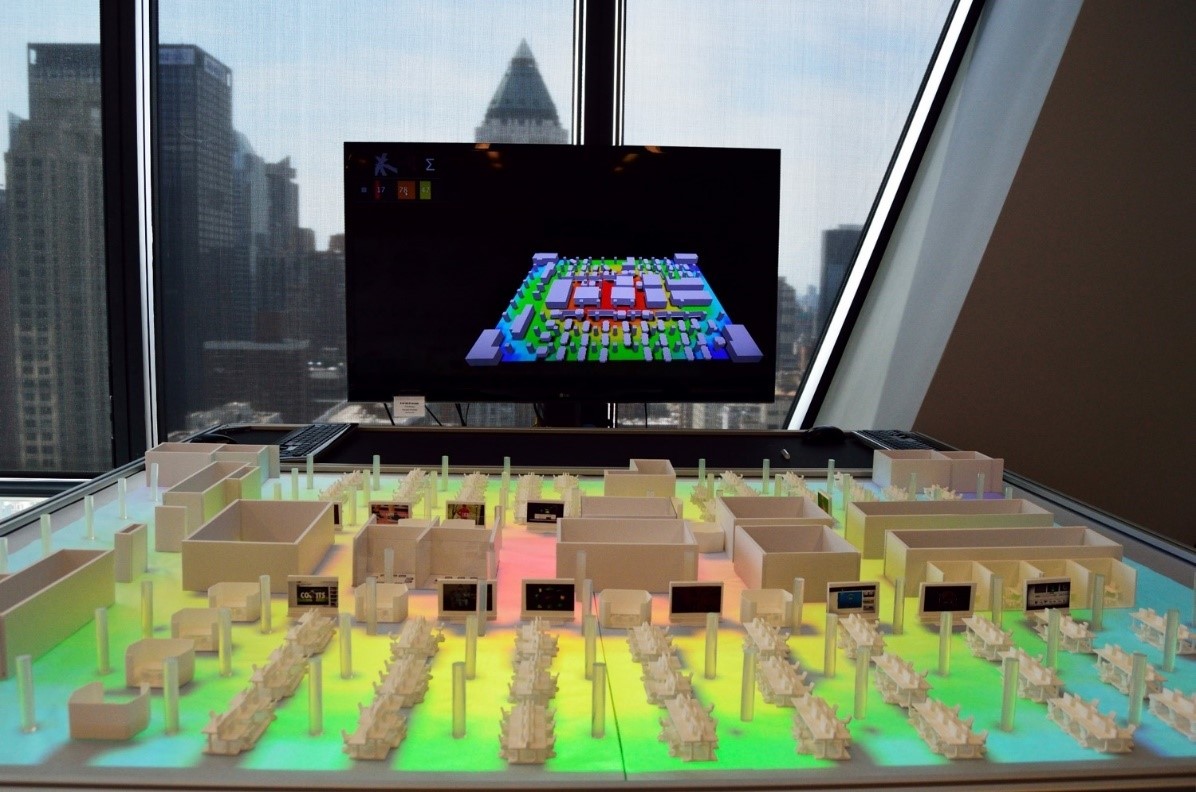}
    \caption{Interactive Physical Modelling environment incorporating visual and spatial connectivity.}
    \label{fig:interactive}
\end{figure}

To facilitate a quick understanding of spatial and visual connectivity of spatial configurations (both in 2D and 2.5D), the authors have developed an array of interactive design tools (one of them portrayed in Figure \ref{fig:interactive}) aiming to provide real-time feedback for the above metrics. Albeit successful to an extent, this has proven quite challenging, particularly when the size of the floor plans was large and the required analysis resolution was high. That is because evaluation of spatial connectivity using Dijkstra’s algorithm \cite{Dijkstra:1959:NTP:2722880.2722945}, as well as visual connectivity using Visibility Graph Analysis \cite{doi:10.1068/b2684} linked to pedestrian movement distribution \cite{Varoudis2015VisibilityAA} can be particularly computationally intensive. Connectivity and VGA analysis could take hours to compute for large scale floor plans. For that reason, many people have attempted to optimize the algorithms used to calculate those metrics, by changing the connectivity's graph creation method \cite{Varoudis2013SPACESA} or by using up-to-date algorithms that utilise multi-core architectures of modern graphics processor devices to do the calculations.\cite{McElhinney} 
Additionally, significant pre and post processing is required in order to incorporate these simulations in interactive applications. This is in order to visualise arbitrary floor plans, while ensuring visual accuracy. This has been further explored by König and Varoudis \citep{20.500.11850/117770} where the workflow was optimized by using interactive visual programming techniques.

Therefore, the main motivation of this research was to explore the use of surrogate models as a replacement for the computationally intensive spatial and visual connectivity simulations (which are used daily within the design workflow), when a real-time outcome is required. To do that, the authors investigated the use of deep neural networks to reduce computation times and resources required to run those simulations substantially.
\subsection{Spatial and visual connectivity}
As mentioned previously, during the design process and especially in the early stages of development, spatial and visual connectivity can be helpful in evaluating the performance of a floor plan. The two analyses closely correlate to ``manifestations of spatial perception, such as way-finding, movement, and space use".\cite{doi:10.1068/b2684}

Spatial connectivity measures the distance to go form any point to any other point on a grid. It, therefore, allows for the analysis of walking distances between points of interest, use cases include analysis of escape routes, analysis of connectivity between different teams within an office plan, etc. \cite{dijkstrasmartexit} \cite{1410938} \cite{articleescapelightcontrol}
It can also be used to understand how isolated (or difficult to reach) a specific space is compared to other spaces within the same floor plan.\cite{10.28945/988}
In contrast, visual connectivity can be used to analyse visual rather than physical distances. Visual connectivity was shown to have a better correlation with real-life observations, this was showcased by Mansouri and Ujang\citep{doi:10.1080/17549175.2016.1213309} while investigating how people are moving within a space and by Shpuza\citep{Shpuza2006FloorplateSA} while studying how well connected teams are within an office. A key concept in the visual connectivity model is that people perceive a given space using their field of view as a guide. This means that during way-finding, people are more likely to navigate within their visual space and even more so in uninterrupted straight lines.\cite{2027}

Both the spatial and visual connectivity analyses require as an input the configuration of the space. This includes the space's internal organization elements, such as walls, partitions and furniture. Particularly, for visual connectivity, the analysis can be run at knee-level, where furniture and walls are included, or at eye-level, where only the walls are included as obstacles.\cite{doi:10.1068/b2684}

In order to run the simulations, the floor plans required pre-processing to be turned into a graph representation. This is achieved by overlaying a grid on the floor plan and using that as the basis of the graph, where each grid cell represents a graph node. For this study, the grid used consisted of square cells, each cell sized at 0.3 meters. It is important to note that in the case of spatial connectivity, the results reported are not substantially affected by the cell's size, as they are reported in metric distance. However, the choice of the cell's size has to meet two conflicting criteria: the cell has to be small enough to retain maximum details about the floor plan, while also being large enough to minimize the number of nodes representing the floor plan for computation time concerns. During the conversion process, some of the nodes in the graph are rendered inaccessible to other graph nodes. This happens when a node is on top of a wall or outside the floor plan boundary. At the end of the pre-processing step, the graph is encoded as a binary image. Nodes in the graph are either represented as black pixels or white pixels. Black indicates obstacles (walls and furniture) or inaccessible nodes, while white indicates accessible open space or accessible nodes. This binary image becomes the input for the next step.

After the nodes are created, the connectivity of the graph is determined. This step differs between the two analyses, as each of them uses different adjacency rules, these rules determine the connections between the different grid cells and they are quite simple. 
For the spatial connectivity, each cell's node is connected with its immediate neighbours, excluding the nodes that are on walls or outside the floor plan boundary. This creates a graph where each node has at most 8 neighbours (4 across and 4 diagonal). Identifying the connections can be easily computed by a convolution over the grid and using a small 3$\times$3 kernel.  

As for visual connectivity, the connections are slightly more complicated, but can be determined by using the following rule, ``two nodes are connected to each other if you can draw a line without crossing an obstacle". There are various methods to determine these connections. The authors used recursive shadow-casting. This is a technique which is commonly used in video games ---where speed is paramount--- to determine the line of sight.\cite{doi:10.1068/b2684} \cite{schaffranekspace} \cite{Turner99makingisovists} 
Recursive shadow-casting was chosen over other slightly more accurate ray-casting techniques, due to its superior speed, which allows for a single pass over the grid resulting in computed visibility. Graph nodes in visual connectivity analysis, can have any number of connections, as opposed to the spatial connectivity's maximum of eight. 

\begin{figure}
    \centering
    \includegraphics[width=\linewidth]{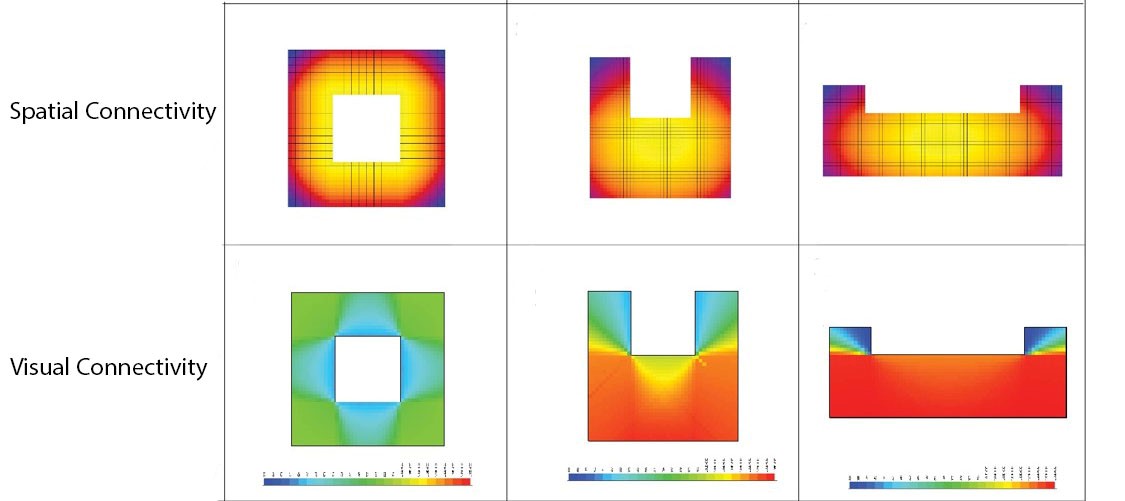}
    \caption{Visual and Spatial connectivity analyses.}
    \label{fig:analysis}
\end{figure}

The product of both analyses is a coloured map that describes areas with high or low connectivity and visibility (see Figure \ref{fig:analysis} for a comparative example). For spatial connectivity, the colors describe the average shortest distance taken to walk from every location to every other location. To calculate that on the graph, Dijkstra's algorithm is used along with a shortest path algorithm.\cite{Dijkstra:1959:NTP:2722880.2722945}
Whereas, for visual connectivity, the colors represent how visually connected a person in a given location would be to all other locations. To calculate this measure, Dijkstra's algorithm is again used to traverse the graph, while the calculations are done according to VGA as described in A. Turner's paper.\citep{doi:10.1068/b2684}

\subsection{Pipeline}
Given that the authors had already developed a simulation engine that could run spatial and visual connectivity analysis, the rest of the process focused on the following aspects: the development of a properly tailored dataset to train deep neural networks and the investigation of the appropriate architecture of the neural network itself. The paper will thus start by presenting the pipeline that allowed for the parametric creation of the floor plans and the automatic completion of their respective analyses using a High Performance Computing system to parallelize the simulation process. It will then delve into the network architectures investigated and will finally discuss their respective success in training the system.
\section{Convolutional Neural Networks for connectivity analysis}
\subsection{Supervised machine learning and analytical models}
There is an increasing body of work that investigates developing surrogate models to speed-up computationally demanding analysis such as computational fluid dynamics analysis \cite{Dijkstra:1959:NTP:2722880.2722945}, finite element analysis, solar radiation etc. Recent advancements in the fields of Machine Learning (ML) and cognitive computing \cite{Jordan255} proved more effective for building such models. The task of performing visual graph or spatial connectivity analysis on a given plan could be formulated in the language of ML as a supervised learning problem.\cite{PatternRecognitionBishop} 
It can be expressed as a mapping between a floor plan with walls and furniture (an input image) to an analysed plan (an output image). The output is some translation of the input, where certain pixels of the image stay unchanged (walls), whereas the others are assigned a value according to the analysis. Those values could then be represented using a color gradient. Therefore, a basic input image is less complex than an analysed image. This task is more common to image processing tasks such as style transfer or colourization than to vision problems like semantic segmentation. In recent years, Convolutional Neural Networks (CNNs) have become popular for a variety of image tasks, because they outperformed previous state of the art techniques.\cite{Girshick} \cite{NIPS2012_4824}
\subsection{Data generation, representation and augmentation}
When machine learning techniques are applied to a new domain, there is usually a deficit of high-quality training data.  Therefore, data collection and labelling become a critical bottleneck. There are different approaches for acquiring data to use for training models, from data discovery to data augmentation and data generation, all three discussed in details by Roh et al.\cite{2018arXiv181103402R} 
In the case of visual and spatial connectivity the input to the system is effectively a 2D plan, but there are no public datasets available to use or to benchmark the models being investigated against. Even though an abundance of data representing 2D plans is publicly available online in an image format, it is not readily usable as the images may contain a lot of noise and are drawn using different styles. This makes them unsuitable in their ``raw" form and would require a lot of laborious preprocessing before they could be used as inputs to any solver. 

\begin{figure} [H]
\centering
\begin{subfigure}[b]{.49\linewidth}
\includegraphics[width=\linewidth]{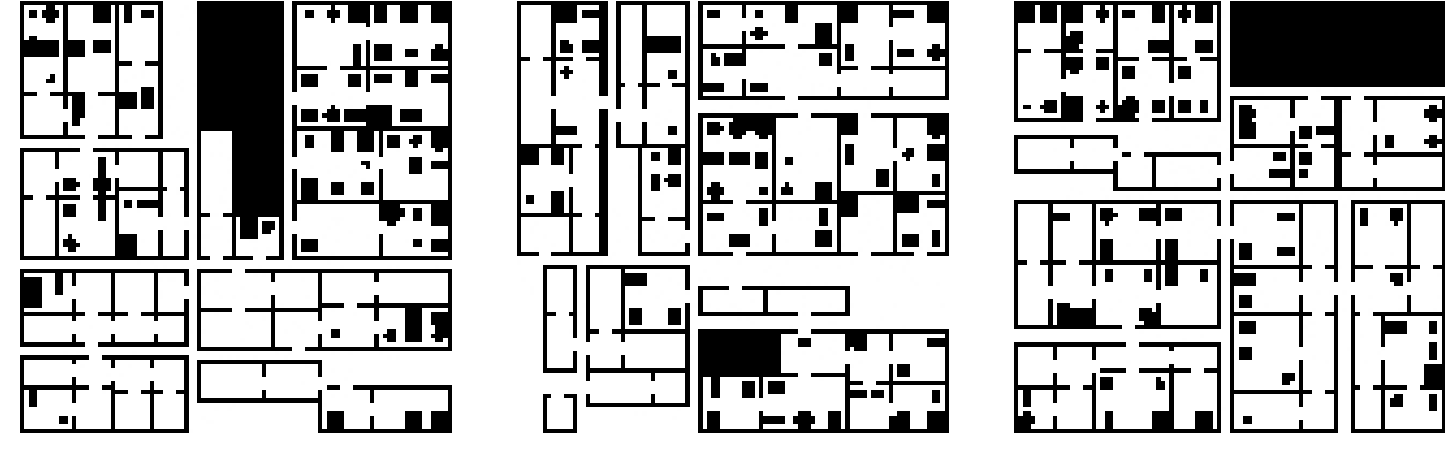}
\end{subfigure}
\begin{subfigure}[b]{.49\linewidth}
\includegraphics[width=\linewidth]{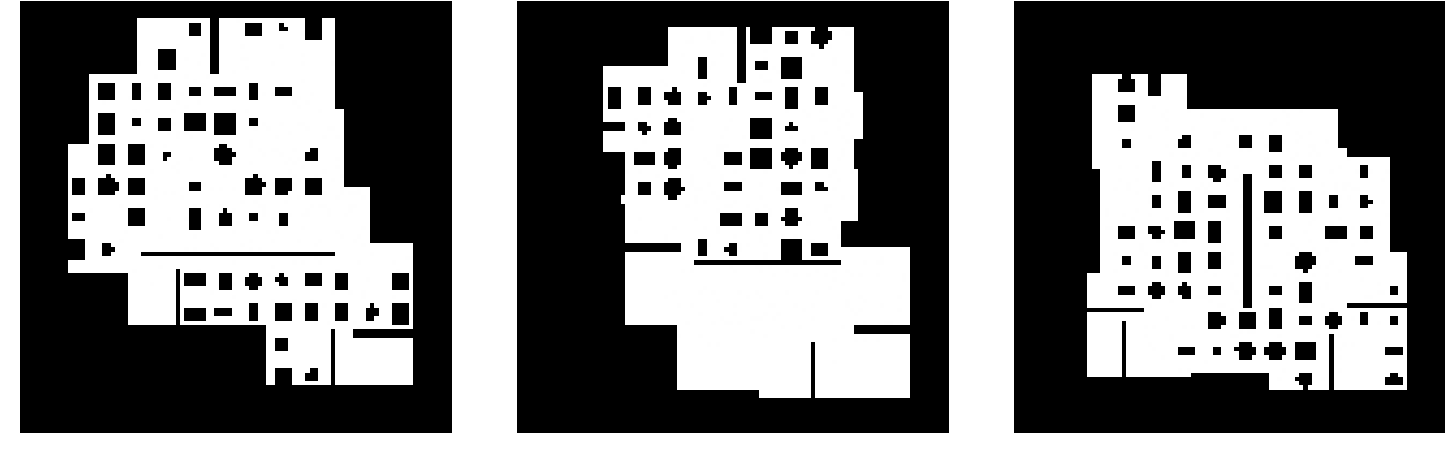}
\end{subfigure}
\begin{subfigure}[b]{.49\linewidth}
\includegraphics[width=\linewidth]{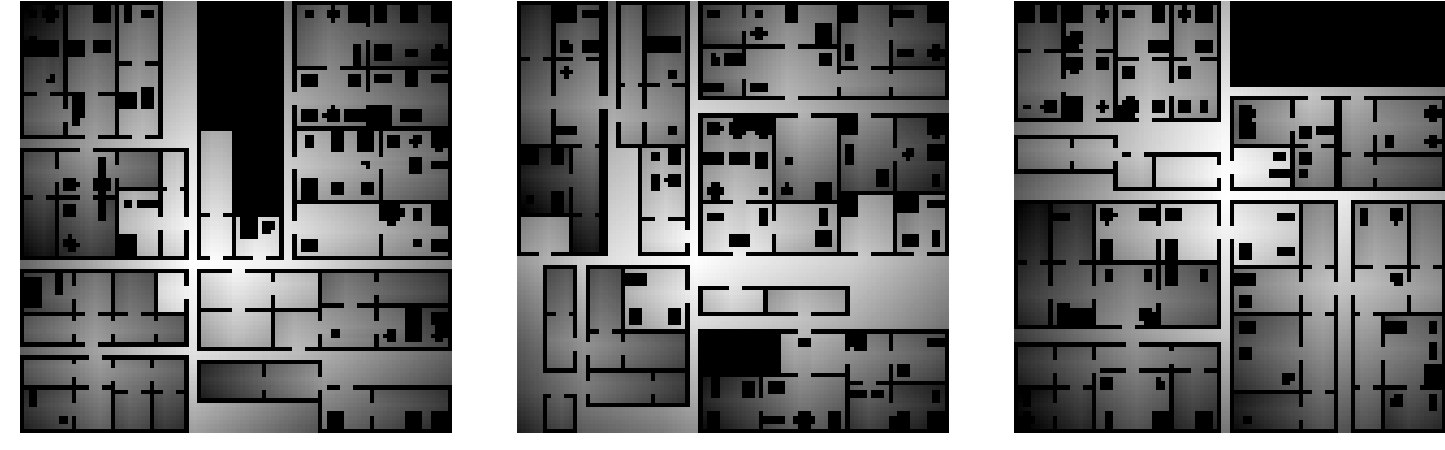}
\end{subfigure}
\begin{subfigure}[b]{.49\linewidth}
\includegraphics[width=\linewidth]{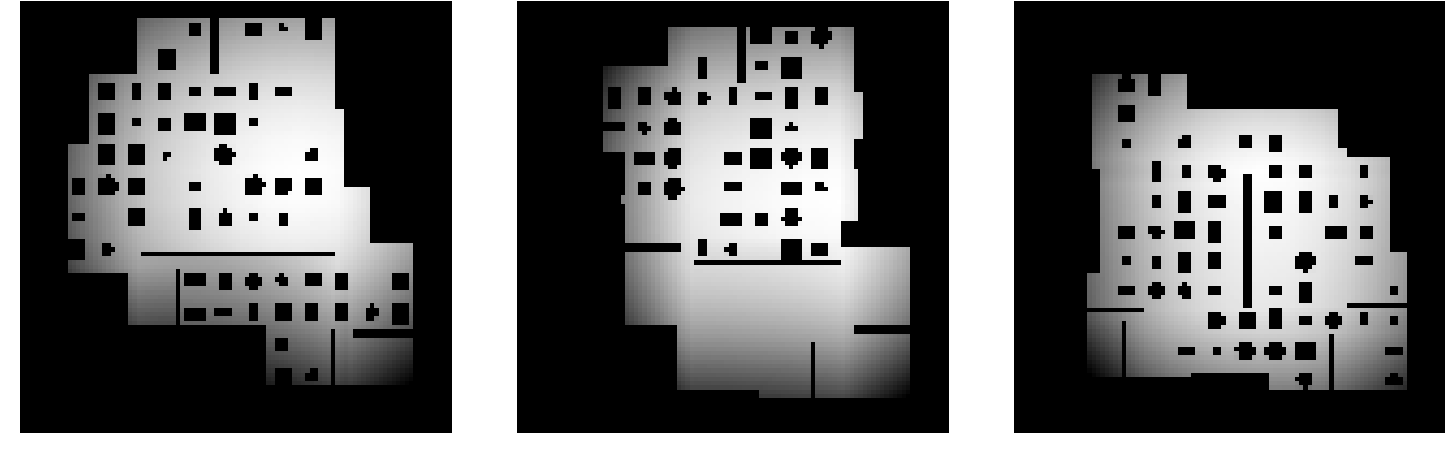}
\end{subfigure}
\begin{subfigure}[b]{.49\linewidth}
\includegraphics[width=\linewidth]{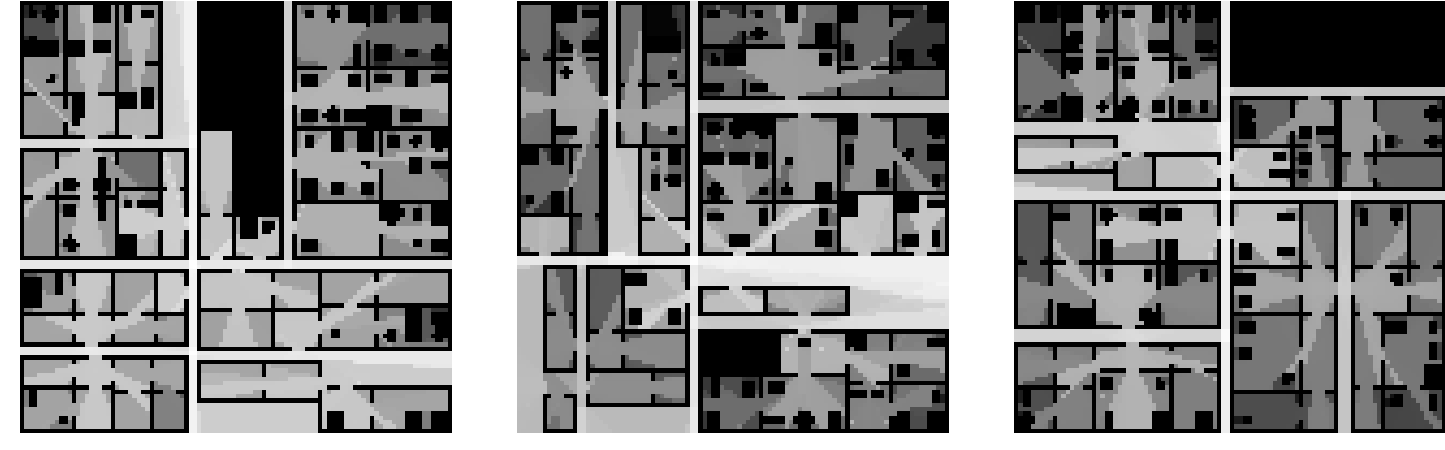}
\end{subfigure}
\begin{subfigure}[b]{.49\linewidth}
\includegraphics[width=\linewidth]{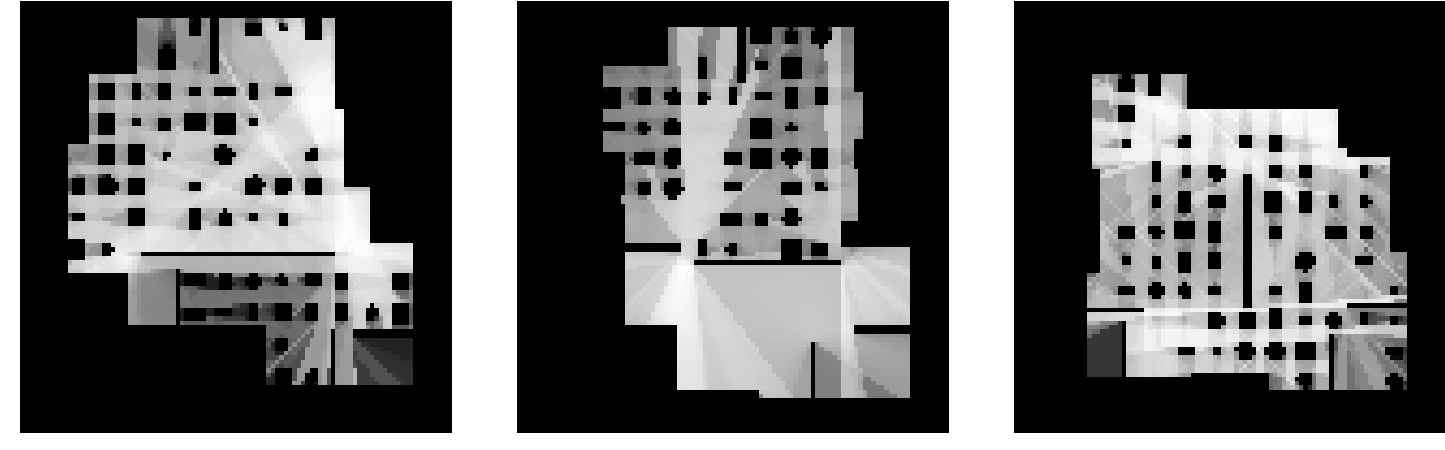}
\end{subfigure}
\caption{Dataset generation and analysis output. First row represents the dataset (binary), as produced via the generative process and representing both compartmentalized and open-plan spaces. Second row represents Spatial Connectivity (0-255) and third row represents Visual Connectivity (0-255).}
\label{fig:dataset}
\end{figure}

Due to the above, a choice was made to focus on synthetic data generation using an automated system. Given the size of the dataset required and how time intensive the spatial and visual connectivity analyses are, modest floor plan sizes were used that would only take a few minutes to compute. A parametric model was implemented in a CAD framework (Rhino and Grasshopper), capable of procedurally generating 2D plans. The plans contained different wall and furniture arrangements, creating a plethora of spatial configurations – both open plan and compartmentalized. Using this parametric model, 6,000 images of different layouts were generated to be used for initial testing, (Figure \ref{fig:dataset}) shows examples of the produced dataset and its respective analyses.

Since the dataset was generated from scratch, the format of the input data could be tailored. The images, as mentioned previously, were a basic binary representation of the plan, where zeroes (black pixels) indicated obstacles and ones (white pixels) indicated accessible zones. The resolution of the produced image was also carefully considered, as the analysis time increased exponentially based on the image size. For that reason, the images were constrained to 100$\times$100 pixels, where 1 pixel represented 1 meter in real space. 

In addition to the above, a Signed Distance Function (SDF) representation for all the plans was generated. An SDF describes the distance of a given point from the closest boundary, it is used in computer graphics and computer vision for Simultaneous Localization and Mapping (SLAM) algorithms (ex. 3D reconstruction using depth cameras).\cite{newcombe2011kinectfusion} \cite{CamerTrackingBylow} \cite{Canelhas644377} 
SDF is expected to encode local geometry details along with global scene structure. Encoding this method of representation along with the binary representation of borders, improved the performance of other ``reduced order models” developed for CFD simulation.\cite{Guo:2016:CNN:2939672.2939738}

Furthermore, to ensure that the previously described dataset was rich enough and to prevent the trained models from memorizing the data and over-fitting, we used on-the-fly augmentation by randomly flipping the images vertically and horizontally during the training process.

The data generated was split into a 70-20-10 training, validation and test split. The test split was later used to evaluate different ML models' performance.
\subsection{Analysing Data}
For the generated images to be analysed based on the bespoke simulation engine developed, a connectivity graph was generated, as discussed in previous chapters. This graph was effectively the actual input to the simulation engine. To expedite the analysis, the graph was reduced by removing all the non-connected nodes (previously black pixels) to make the calculation quicker. Furthermore, any isolated sub-graph within the network was disregarded. After the original (pre-analysis) dataset was processed based on the above rules, each image was analysed by the custom spatial and visual connectivity simulation engines, which were abstracted to run in a headless mode, using a simple command line argument. These algorithms were highly parallelized, so the code was adjusted to take advantage of all the cores of the system.
Normally, running these analyses would require high computational resources, and it would result in long run times. This could become a significant bottleneck given the amount of simulations that needed to be run to produce a big enough training set for a CNN. In computing, such problems are tackled by using massive parallelization. Nevertheless, this concept has not been widely used in architecture. However, there are some notable examples which include a prototype of an evolutionary optimization system, which could distribute phenotype and fitness calculations using Microsoft Azure cloud computing system\cite{inproceedings}; utilization of a shared network system to speed up daylight autonomy simulation using Radiance\cite{207b3b0306194391ac90c9f749a9cb5a}; distribution of Radiance based annual daylight simulation over Microsoft Azure.\cite{inproceedings}
In the discussed pipeline this problem was tackled in two ways:  the first was to multi-thread the analysis and take advantage of the full capabilities of the available hardware and the second was to spread all the data over multiple desktop machines and compute them in parallel. These two mitigations essentially reduced the time it would take to produce the training dataset from a month to a few hours.
The overall parallelization process was facilitated by using High Performance Computing system called Hydra. Hydra\cite{hydraaa} is a bespoke in-house developed system, built on top of an on-premise render farm and it was used to distribute simulation tasks to 16 concurrent computing nodes. Each node used an Intel Xenon CPU E5-2660 processor. This allowed for a massive performance speed up. For a detailed breakdown please refer to (Table \ref{table:Computationstats}).

The output (post-analysis dataset) was a set of 100$\times$100 pixel images, with each pixel having its value remapped in grey-scale from 0 to 255, based on the analysis results.
\begin{table}[h]
\small\sf\centering
\caption{Computation statistics of analysis speed-up using Hydra.\label{table:Computationstats}}
\begin{tabular}{lll}
\toprule
Dataset&Corridors&Corridors\\
\midrule
\multicolumn{1}{p{4cm}}{\raggedright Analysis}&VGA&Con\\
\multicolumn{1}{p{4cm}}{\raggedright No. of samples}&3002&3002\\
\multicolumn{1}{p{4cm}}{\raggedright Total CPU time $[dd:hh:mm:ss]$}&02:07:50:53&00:14:36:15\\
\multicolumn{1}{p{4cm}}{\raggedright Actual Evaluation Time $[hh:mm:ss]$}&06:00:57&02:47:09\\
\multicolumn{1}{p{4cm}}{\raggedright Speed-up}&9.16&5.25\\
\bottomrule
\end{tabular}\\[10pt]
\begin{tabular}{lll}
\toprule
Dataset&Open Plan&Open Plan\\
\midrule
\multicolumn{1}{p{4cm}}{\raggedright Analysis}&VGA&Con\\
\multicolumn{1}{p{4cm}}{\raggedright No. of samples}&3002&3002\\
\multicolumn{1}{p{4cm}}{\raggedright Total CPU time $[dd:hh:mm:ss]$}&02:19:41:07&00:08:57:31\\
\multicolumn{1}{p{4cm}}{\raggedright Actual Evaluation Time $[hh:mm:ss]$}&06:40:33&02:21:24\\
\multicolumn{1}{p{4cm}}{\raggedright Speed-up}&10.14&3.8\\
\bottomrule
\end{tabular}\\[10pt]
\end{table}
\subsection{Network architecture and implementation}
While various possible architectures for the neural network were investigated, the main requirement was for the Convolutional Neural Network (CNNs) to be able to scale up. That meant that the architecture of the CNN system needed to allow the input to be of arbitrary size and not restricted to the input size on which the network would be trained. This led to the use of a Fully Convolutional Network (FCN).\cite{Shelhamer:2017:FCN:3069214.3069246} 
Another concern was the need for an architecture that would offer good localization and manage to propagate context information through-out the whole model, which pointed at the use of a U-Net network.\cite{2015arXiv150504597R} 
The U-Net architecture is symmetrical and contains two parts: a contracting path and   an expansive path, which gives it the `U' shape. The number of operations along the two paths is the same. 

The contracting part is a typical CNN, and is made of successive down-sampling operation blocks, successively applied on top of each other, with the first applied to the input image. The objective is to ensure that the network effectively learns complex structures and features. Each operational block consists of two 3$\times$3 convolution layer, followed by a 2$\times$2 max pooling operation. The purpose of those steps is to reduce the dimensionality of the input while increasing the number of features extracted. The number of trainable filters/feature-detectors after each down-sampling operation is doubled (64, 128, 256, 512 and 1024), hence so does the resulting feature maps. A feature map is the result of convolving with a filter/feature-detector over the input image or the output of a previous operation block.

The expansion part is symmetrical to the contraction one but contains only up-sampling blocks, trying to reconstruct the image from the output of the last down-sampling block. Each up-sampling block passes the input through two 3$\times$3 convolution layers then a 2$\times$2 up-sampling layer. To maintain symmetry with the contracting part, the number of of trainable filters/feature-detectors gets halved in each new block (1024, 512, 256, 128, 64). To preserve the contextual/spatial information and the features learned in the down-sampling path, the input to each successive expansion block is augmented by the feature maps from the corresponding contraction block, this ``knowledge transfer" between the two paths is called skip connections. More details about the model's architecture can be seen in Ronneberger et al. work.\cite{2015arXiv150504597R} The model's implementation was done using Tensorflow in Python.

Some adjustments were introduced to the U-Net model, while the original use of the model was to do per-pixel classification, the final activation layer was changed to a sigmoid function to accommodate the task in hand, which can be described as a per-pixel regression task. Different optimization algorithms were tested first on a subset of the data, while using different learning rates.

\subsection{Training}
When training a machine learning model, there are some hyper-parameters that require tuning. Two of the prominent parameters to tune are the learning rate and the choice of the optimization algorithm. Using the U-Net model, different optimization algorithms were tested. Those included Stochastic Gradient Decent, Adam, RMSProp and Adadelta.\cite{2012arXiv1212.5701Z}
The tests took place on a small subset of 50 images from the dataset and the best performers were SGD and Adadelta. The models using these optimizers converged quickly, managing at the beginning of the training to map black pixels in the input to black pixels in the output, while the rest - on more than one occasion - were stuck fading the black pixels and replacing them with grey or white. Therefore, SGD and Adadelta were selected for the experiments to follow. With both optimizers, the consequent training was initiated with different learning rates and an increased subset of 100 images. Between the two optimizers and during different runs, Adadelta managed to converge faster. 

Besides tracking the training loss, the model was also tested in each step on a fixed number of images, so that the training progress could be observed and traced. This process allowed visualizing the effect of the optimizers and the chosen learning rates. In addition, it provided a visual and intuitive way of evaluating early on if the model was converging towards a desired minima or not.

After settling on those two parameters (Adadelta as an optimizer and a learning rate of 1.0) the model was trained on the whole training set using Mean Squared Error (MSE) as the loss function. Loss functions or cost functions are an essential component in a neural network, helping and guiding the optimization algorithm explore the solution space. This function guides the network to make desired results, for example to create sharp edges in the context of VGA. The goal of the function is, given an analysis prediction from the model to an input plan, and knowing the correct output of the analysis for that plan, calculate how much the model's prediction deviated from the correct answer. Loss functions design is considered a pressing problem which is heavily researched and requires expert knowledge. In this case, using MSE, the loss was decreasing, but when comparing the model's prediction and the actual analysis side by side, it was revealed that the distribution of values was not visually matching. Using other simplistic loss functions like Absolute Error or Mean Square Logarithmic Error did not help. A more tailored loss function seemed to be necessary. That led to experimenting with image Gradient Difference Loss (GDL) developed by Mathieu et al. \cite{2015arXiv151105440M} which used a weighted sum along with MSE. GDL takes into consideration the neighbouring pixels intensity differences. This enhanced the performance of the model, but still didn't reach the targeted accuracy. It is worth noting at this point that the use of a Signed Distance Function representation of the plans, did not increase the performance of the network, so only the binary representation was used as an input.

It was then decided to experiment with a different model architecture based on Generative Adversarial Networks (GAN) where two models, a generator and a discriminator, are competing to accomplish a defined task. This architecture depends partially on the U-Net architecture used before, but instead of using explicit loss functions, the discriminator network is trying to learn a loss function suitable for the task in hand. This architecture basically transforms the loss function to a trainable parameter in the model. The two networks then compete: the generator network tries to learn how to generate convincing images against its adversary, the discriminator, which tries to judge whether an image presented to it is original (from the training data) or synthesized by the first network. 

This architecture avoids hand-engineering of the loss function and incentivizes the network to produce images which could be undistinguishable from reality. An effective implementation of such an architecture is Pix2Pix, a conditional Generative Adversarial Network (cGAN) developed by Isola et al.\cite{2016arXiv161107004I}
It was designed as a general-purpose image-to-image translator, that can be conditioned to generate a specific translation given an input image, instead of just a randomly generated image.  In this approach the generator network is presented with both an input image and a random noise vector and learns to produce images that the discriminator cannot tell apart from the genuine analyses output.  The Pix2Pix generator network uses the encoder-decoder pattern with skip connections form the U-Net architecture, to preserve low-level information shared between the input and output. 

\begin{figure}
\centering
\begin{subfigure}[b]{0.99\linewidth}
\includegraphics[width=\linewidth]{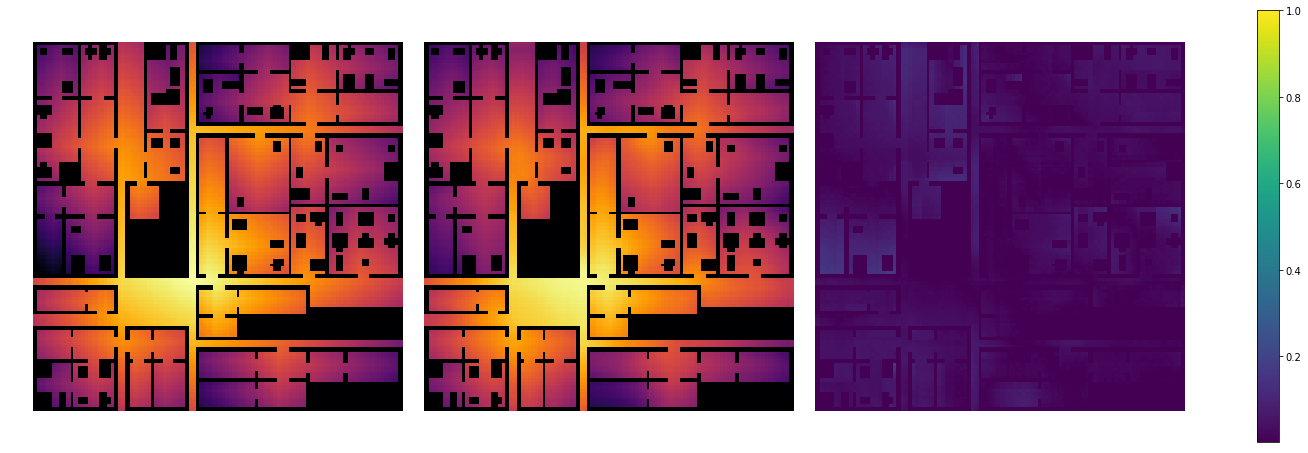}
\caption{Minimum loss scored in the test set using U-Net.}
\end{subfigure}
\begin{subfigure}[b]{0.99\linewidth}
\includegraphics[width=\linewidth]{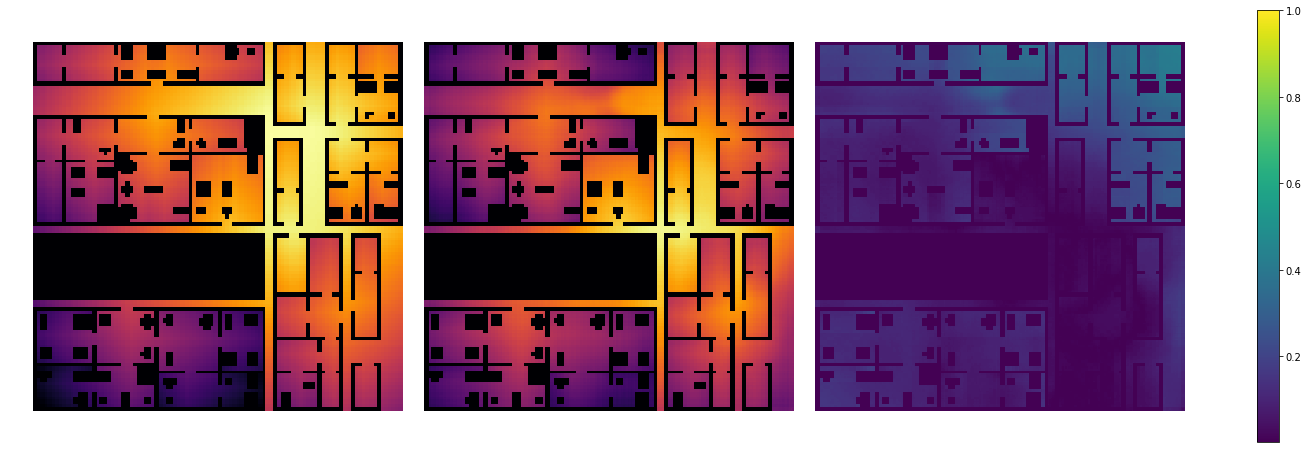}
\caption{Maximum loss scored in the test set using U-Net.}
\end{subfigure}
\begin{subfigure}[b]{0.99\linewidth}
\includegraphics[width=\linewidth]{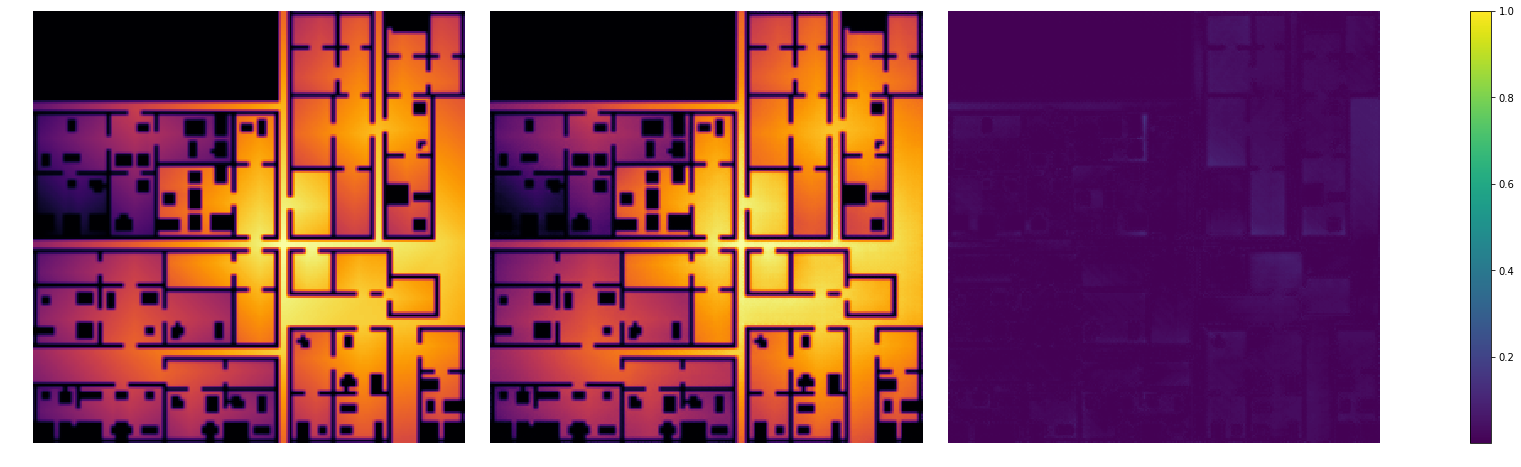}
\caption{Minimum loss scored in the test set using Pix2Pix.}
\end{subfigure}
\begin{subfigure}[b]{0.99\linewidth}
\includegraphics[width=\linewidth]{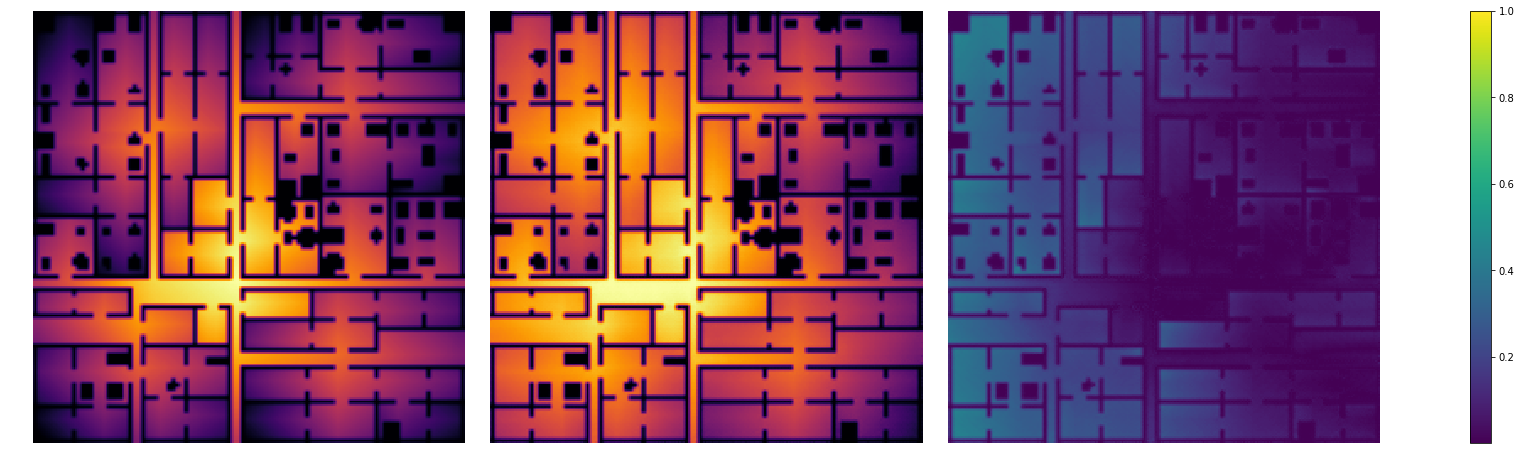}
\caption{Maximum loss scored in the test set using Pix2Pix.}
\end{subfigure}
\caption{Visual comparison between the images that the models scored lowest and highest loss value for. In each, the left image is the expected output, the middle shows the model's predicted output and the last image to the right shows the delta between both, darker means closer to the correct answer.}
\label{fig:learningOut}
\end{figure}

\begin{figure}
\centering
\begin{subfigure}[b]{0.99\linewidth}
\includegraphics[width=\linewidth]{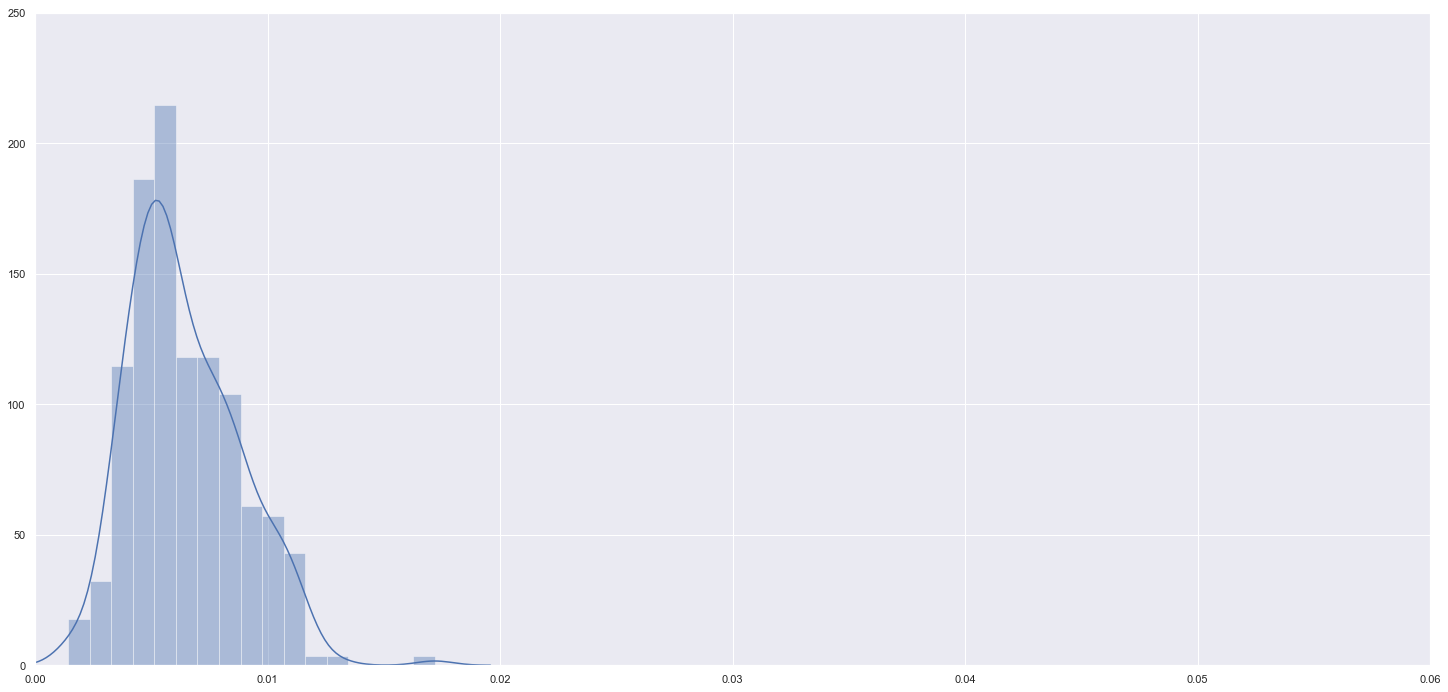}
\caption{UNet loss distribution for test set.}
\end{subfigure}
\begin{subfigure}[b]{0.99\linewidth}
\includegraphics[width=\linewidth]{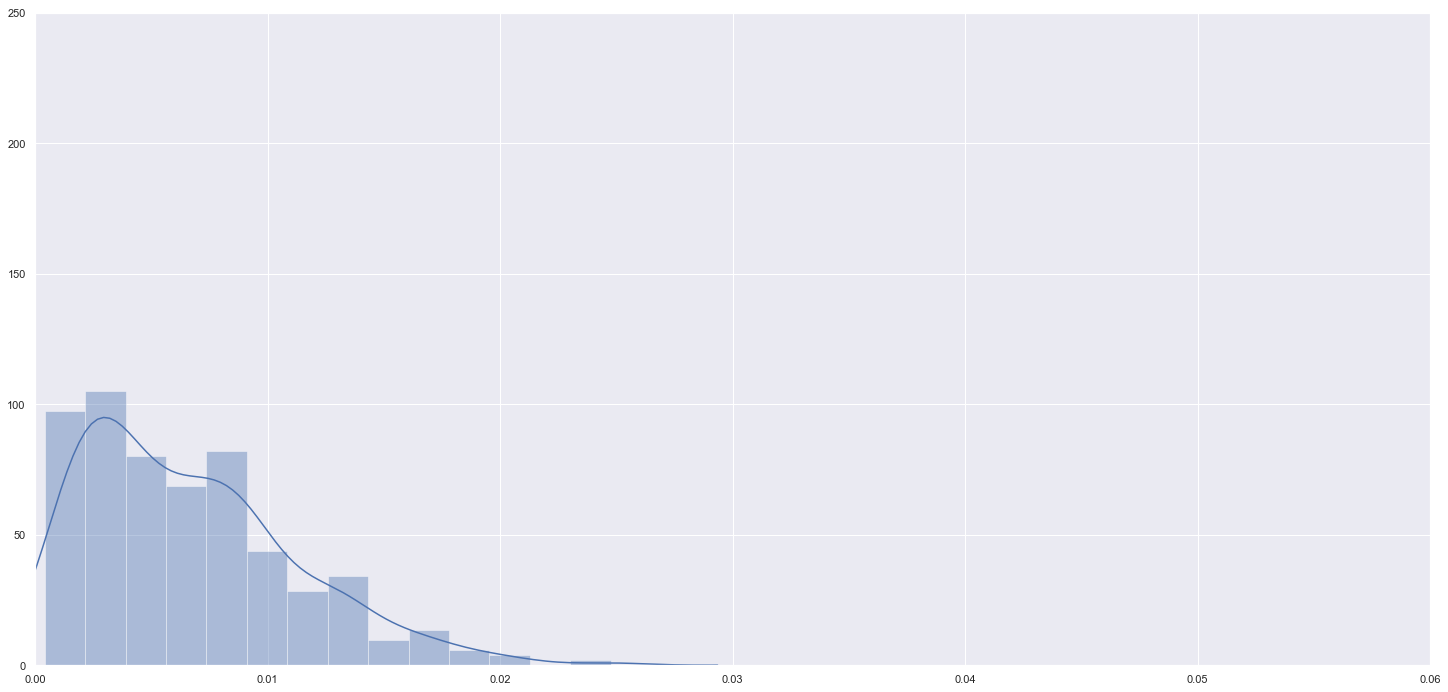}
\caption{Pix2Pix loss distribution for test set.}
\end{subfigure}
\caption{distribution of losses for U-Net, Pix2Pix generating Connectivity Analysis and Pix2Pix generating VGA.}
\label{fig:lossDist}
\end{figure}

Table \ref{table:learning} compares the performance of the two networks: the fully convolutional U-Net network and the Conditioned Generative Adversarial Pix2Pix network, both trained on the same data subset, while Figure \ref{fig:learningOut} showcase the visual results. For each sub-figure the picture to the left shows the analytical result while the one in the middle shows the network-generated analysis. The third picture always demonstrates the delta between the two (darker parts mean the more optimal the results output by the network). Finally, Figure \ref{fig:lossDist} demonstrates the distribution of losses for U-Net, Pix2Pix generating Connectivity Analysis.
\begin{table}
\small\sf\centering
\caption{Model Loss Comparison between U-Net network and the Conditioned Generative Adversarial Pix2Pix network.\label{table:learning}}
\begin{tabular}{lll}
\toprule
Model&U-Net&Pix2Pix\\
\midrule
\multicolumn{1}{p{4cm}}{\raggedright MSE Loss $(300 images test set)$}&0.0064&0.0066\\
\multicolumn{1}{p{4cm}}{\raggedright minimum}&0.00148&0.00041\\
\multicolumn{1}{p{4cm}}{\raggedright maximum}&0.0167&0.0248\\
\bottomrule
\end{tabular}\\[10pt]
\end{table}

\section{Discussion}
One of the goals of training a neural network compared to running the actual analysis, was to optimize the speed of the outcome. This is especially useful as the time it takes to run one image is not dependent on the floor plan scale or the type of analysis (as in the actual analytical models), but on the size of the image and the type of network used. For the Pix2Pix architecture, inference (predicting the output given an input image) for one image is computed in 0.08 seconds and for the U-Net in 0.032 seconds for each of the analyses, compared to 15 seconds for running the actual spatial connectivity analysis and 128 seconds for running the visual connectivity. It is also important to note, that for running the analysis a 100$\times$100 pixels image was used and for training the dataset a 256$\times$256 pixels image, so the data was scaled up. This means that there is a potential of using the same network architecture for more detailed and/or larger floor plans.

Since Pix2Pix tries to learn a loss function well suited for the problem in hand, it is easier to start experimenting with, there is no need to tweak anything in the model’s architecture and hyper-parameters. This is useful for quickly testing whether an idea, or a collected dataset is appropriate. Choosing a suitable loss function in UNet and tweaking other hyper-parameters, requires more time and experience, and in this case, led to better performance. Moving from simple loss functions, like Mean Square Errors or Absolute Errors to more sophisticated ones like image Gradient Difference Loss, helped increase the performance of a fully convolutional network like U-Net.

While evaluating a model's performance, it is important to study the distribution of losses. Even though Pix2Pix reported lower loss on the test data set, plotting the losses for both models as seen in Figure \ref{fig:lossDist} shows that the variance in UNet is lower. This means that the UNet model manages to generalize better on almost all the test examples.

Even though training the model requires a lot of time, this can be resolved by using newer GPUs optimized for machine learning tasks. This can be seen in the speed improvements in training time in Table \ref{table:traintime}. The UNet model had 31,031,685 trainable parameters in total. The training set was composed of 2,100 images of size 256$\times$256 pixels, each just one channel. The batch size (number of images the model gets to try and predict the output for, before the cost function is calculated and the parameters of the model are updated by the optimizer) was 8 images. The experiments ran for around 500 epochs (one epoch is 262 batches or the length of the dataset by the batch size). For the comparison, the average time it took to train per batch was taken into consideration. Each batch included feeding forward calculations for the total number of images per batch, calculating loss, averaging it and then doing back propagation. The number of images per batch was being tweaked and hence varied during some of the experiments. So the total computation time was divided by the number of images which, albeit unusual, made the numbers comparable. Effectively, based on the numbers in Table \ref{table:traintime}, the acceleration compared to the baseline Nvidia's P4000 card that was used, was 66\% using Nvidia's RTX4000 card and 100\% using Nvidia's RTX6000 card.

\begin{table}
\small\sf\centering
\caption{Time per image during training the UNet model on different NVidia GPUs.\label{table:traintime}}
\begin{tabular}{llll}
\toprule
Model&P4000&RTX4000&RTX6000\\
\midrule
\multicolumn{1}{p{3cm}}{\raggedright Training time $(sec/image)$}&0.714&0.429&0.357\\
\bottomrule
\end{tabular}\\[10pt]
\end{table}

\section{Future work}
Although further steps need to be taken to improve the reporting of the metrics, the current speed and accuracy provide some interesting possibilities for further development. Networks such as the ones presented here could be used for the creation of interactive applications that provide real-time feedback to designers as they design a space. Up until now, in order to achieve that, one should sacrifice either accuracy or time. This doesn't have to be the case, particularly now, as due to the popularity of Machine Learning, GPU manufacturers currently provide highly optimized hardware for running and training neural networks, while also making it accessible and cheaper. 

It would also be worthwhile investigating further augmentation techniques, especially if the advantages of U-Net are to be utilized, as well as going through a more exhaustive testing of both networks on corner cases. We leave these and other improvements for future work.

The results of this process have been encouraging enough to indicate potential expansion of this system into an array of different analysis that require a floor plan as an input - particularly environmental analysis like glare, that seems to be a recurring problem in the architectural sector. 

Another point of interest is looking into means of converting images of plans into a unified queryable representation. This would aid in bypassing the laborious pre-processing required to clean up the images, making them feasible to be used as a training dataset. In addition to potentially providing access to a huge number of floor plan images publicly available online, instead of creating synthetic datasets.

\section{Conclusion}
This paper presented the process of using deep learning to train a surrogate model that could output spatial and visual connectivity for any given floor plan in real-time. The use of spatial and visual connectivity in architecture was discussed, as well as the computationally intensive requirements of both these algorithms. An implementation of a Fully Convolutional Network was then presented as a potential deep learning model to replace the analytical models. It was also compared to another more recent model architecture which is the Generative Adversarial Networks architecture. To that end, the authors showcased how the training set was generated and analysed for the supervised machine learning process, and how the network architecture, implementation and training were conducted. The predicted analyses output from our trained model are orders of magnitude faster compared to usual calculation approaches, at the cost of minimal error rates. The paper concluded with the ongoing results of the research and the next steps that would be required for its continuation.

\bibliography{./Thesis}

\end{document}